\documentclass[10pt,twocolumn,letterpaper]{article}

\usepackage{cvpr}
\usepackage{times}
\usepackage{epsfig}
\usepackage{graphicx}
\usepackage{amsmath}
\usepackage{amssymb}
\usepackage{array}
\usepackage[ruled,vlined,commentsnumbered]{algorithm2e}
\usepackage{url}
\usepackage{dsfont}

\cvprfinalcopy 

\def\cvprPaperID{****} 

\begin{document}

\title{Learning a Convolutional Neural Network for Non-uniform \\ Motion Blur Removal}
\author{Jian Sun\textsuperscript{1},  Wenfei Cao\textsuperscript{1},  Zongben Xu\textsuperscript{1},  Jean Ponce\textsuperscript{2,} \thanks{WILLOW project-team, D\'epartement d'Informatique de l'Ecole Normale
Sup\'erieure, ENS/Inria/CNRS UMR 8548.}
\\
\textsuperscript{1}Xi'an Jiaotong University, \ \
\textsuperscript{2}\'{E}cole Normale Sup\'erieure / PSL Research University 
}

\maketitle

\begin{abstract}
In this paper, we address the problem of estimating and removing non-uniform motion blur from a single blurry image. We propose a deep learning approach to predicting the probabilistic distribution of motion blur at the patch level using a convolutional neural network (CNN). We further extend the candidate set of motion kernels predicted by the CNN using carefully designed image rotations. A Markov random field model is then used to infer a dense non-uniform motion blur field enforcing motion smoothness. Finally, motion blur is removed by a non-uniform deblurring model using patch-level image prior. Experimental evaluations show that our approach can effectively estimate and remove complex non-uniform motion blur that is not handled well by previous approaches.
\end{abstract}

\vspace{-0.3cm}
\section{Introduction}
\vspace{-0.1cm}
Image deblurring~\cite{cho2007removing, fergus2006removing,Hu2008blur,   joshi2008psf, krishnan2009fast,  michaeli2014blind,sun2014good, yuan2007image}  aims at recovering sharp image from a blurry image due to camera shake, object motion or out-of-focus. In this paper, we focus on estimating and removing spatially varying motion blur.

\begin{figure*}
\begin{center}
\includegraphics[width=1.0\linewidth]{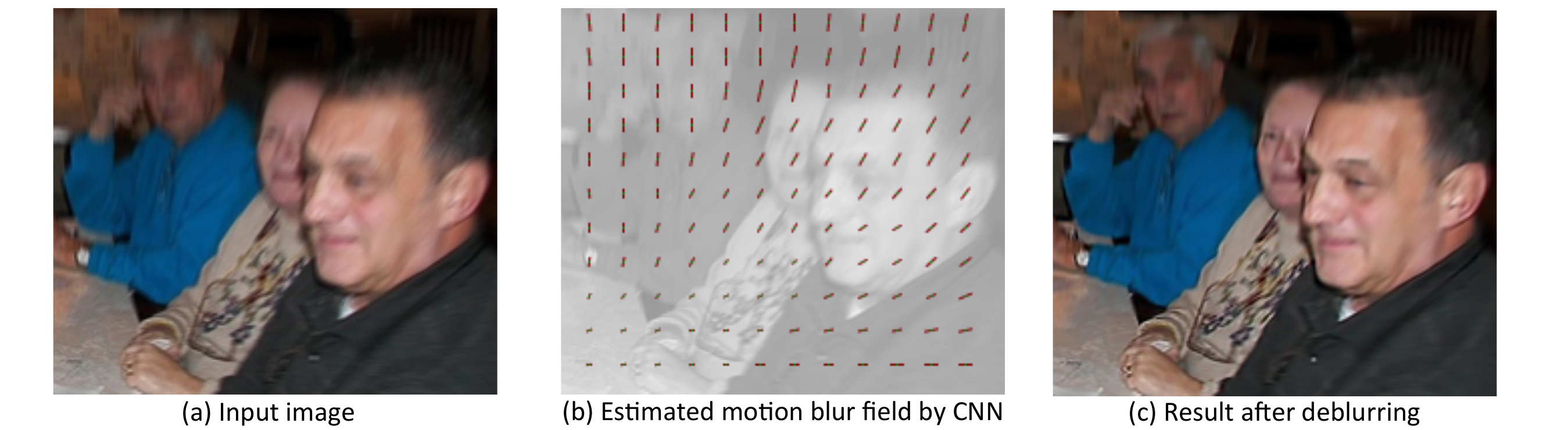}
\end{center}
\vspace{-0.3cm}
\caption{An example illustrating our approach. Given an image with non-uniform motion blur (left). We first estimate the field of non-uniform motion blur kernels by a convolutional neural network (middle), then deconvolve the blurred image (right).}
\label{fig:ex_first}
\vspace{-0.5cm}
\end{figure*}

Non-uniform deblurring~\cite{ji2012motion,kim2014segmentation,paramanand2013non} has attracted much attention in recent years. Methods in~\cite{gupta2010single, hirsch2011fast, whyte2012, zheng2013forward} work on non-uniform blur caused  by camera rotations, in-plane translations or forward out-of-plane translations. They are effective for removing non-uniform blur consistent with these motion assumptions.  Another category of approaches works on non-uniform motion blur caused by object motion. They estimate blur kernels by analyzing image statistics~\cite{levin2007blind}, blur spectrum~\cite{chakrabarti2010analyzing}, or with a learning approach using hand-crafted features~\cite{couzinie2013learning}. Other approaches~\cite{kim2014segmentation,xuL0} jointly estimate the sharp image and blur kernels using a sparsity prior. It is still challenging today to remove strongly non-uniform motion blur captured in complex scenes.





In this work, we propose a novel deep learning-based approach to estimating non-uniform motion blur, followed by a patch statistics-based deblurring model adapted to non-uniform motion blur. We estimate the probabilities of motion kernels at the patch level using a convolutional neural network (CNN)~\cite{donahue2013decaf, kangconvolutional, krizhevsky2012imagenet, LeCun},  then fuse the patch-based estimations into a dense field of motion kernels using a Markov random field (MRF) model. To fully utilize the CNN, we propose to extend the candidate motion kernel set predicted by CNN using an image rotation technique, which significantly boost its performance for motion kernel estimation. Taking advantage of the strong feature learning power of CNNs, we can well predict the challenging non-uniform motion blur that can hardly be well estimated by the state-of-the-art approaches. 

Figure~\ref{fig:ex_first} illustrates our approach. Given a blurry image, we first estimate non-uniform motion blur field by a CNN model, then we deconvolve the blurry image. Our approach can effectively estimate the spatially varying motion kernels, which enable us to well remove the motion blur.



\subsection{Related Work}
Estimating accurate motion blur kernels is essential to non-uniform image deblurring.
In~\cite{gupta2010single,hirsch2011fast,tai2011richardson,whyte2012, zheng2013forward}, non-uniform motion blur is modeled as a global camera motion, which basically estimates an uniform kernel in the camera motion space. Methods in~\cite{ji2012motion,kim2014segmentation,xuL0} jointly estimate the motion kernels and sharp image. They rely on a sparsity prior to infer the latent sharp image for better motion kernel estimation. Different to them, we estimate motion blur kernels directly using the local patches, which does not require the estimation of camera motion or a latent sharp image.


Another category of approaches~\cite{chakrabarti2010analyzing,dai2008motion} estimates spatially varying motion blur based on local image features.
The method in~\cite{chakrabarti2010analyzing} estimates motion blur based on blur spectrum analysis of image patch in Fourier transform space. ~\cite{levin2007blind} predicts motion blur kernel using natural image statistics. \cite{dai2008motion} estimates motion blur by analyzing the alpha maps of image edges. \cite{couzinie2013learning} learns a regression function to predict motion blur kernel based on some hand-crafted features. 
Different to them, we estimate motion blur kernels using a convolutional neural network, followed by a carefully designed  motion kernel extension method and MRF model to predict a dense field of motion kernels. Our approach can well estimate complex and strong motion blur, which can hardly be well estimated by the previous approaches.

Recently, there has been some related work on learning-based deblurring approaches. 
\cite{schmidt2013discriminative} proposes a discriminative deblurring approach using cascade of Gaussian CRF models for uniform blur removal. \cite{schuler2013machine} proposes a neural network approach for learning a denoiser to suppress noises during deconvolution. \cite{xu_nips} designs an image deconvolution neural network for non-blind deconvolution. These approaches above focus on designing better learning-based model for uniform blur removal. Our approach works on a more challenging task of non-uniform motion blur estimation and removal. Our CNN-based approach provides an effective method for solving this problem.


\vspace{-0.14cm}
\section{Learning a CNN for Motion Blur Estimation}

We propose to estimate spatially-varying motion blur kernels using a convolutional neural network. The basic idea is that we first predict the probabilities of different motion kernels for each image patch.  Then we estimate dense motion blur kernels for the whole image using a Markov random field model enforcing motion smoothness. 

\begin{figure}[!htbp] 
\begin{center}
\includegraphics[width=1.0\linewidth]{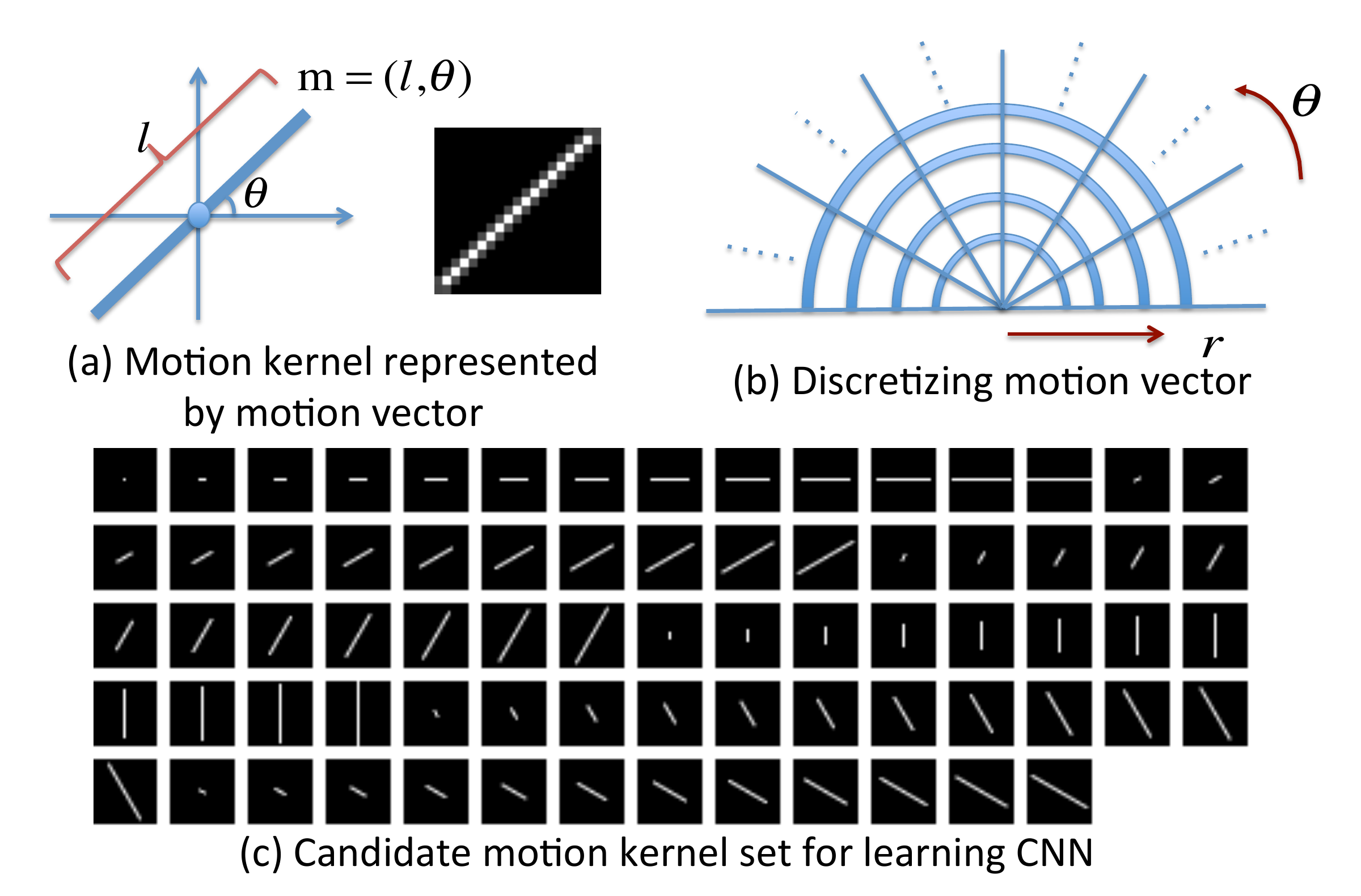}
\vspace{-0.6cm}
\end{center}
   \caption{Representation of motion blur kernel by motion vector and generation of motion kernel candidates.}
\label{fig:motion_discretize}
\vspace{-0.3cm}
\end{figure}

Before giving the details of our approach, let us first introduce our general formulation for non-uniform motion blur. We consider non-uniform image blur caused by object or camera motion.  Given a blurry image $I$, we represent the local motion blur kernel at an image pixel $p \in \Omega$ ($\Omega$ is the image region) by a \textit{\textbf{motion vector}} $\mathbf{m_p} = (l_p, o_p)$, which characterizes the length and orientation of the motion field in $p$ when the camera shutter is open. As shown in Fig.~\ref{fig:motion_discretize}(a), each motion vector determines a motion kernel with  non-zero values only along the motion trace. The blurry image can then be represented by $I = k(M) * I_0$, i.e., the convolution of a latent sharp image $I_0$ with the non-uniform motion blur kernels $ k(M)$ determined by the motion field $M = \{\mathbf{m_p}\}_{p \in \Omega}$. 

In the following paragraph, we also represent the motion vector $\mathbf{m_p} $ as $(u_p, v_p)$ in Cartesian coordinate system based on the transform:
\begin{eqnarray} \label{eqn:catesian}
u_p = l_p  \cos({o_p}), 
v_p = l_p \sin({o_p}).
\end{eqnarray}

The estimation of spatially-varying motion blur kernels is equivalent to estimating the motion field\footnote{Note that motions $\mathbf{m} = (l, o)$ and $\mathbf{m}' = (l, o + 180^{\circ})$ generate the same motion blur kernel. We therefore only need to estimate the motions with $o \in [0, 180^{\circ})$.} from a single blurry image. In our approach, we do not make any global parametric assumptions (e.g., homography) on the motion, therefore the motion kernel estimation is challenging, and we only use local image regions for predicting these kernels.



\subsection{Patch-level Motion Kernel Estimation by CNN} \label{sec:cnnMotion}

\begin{figure*}[htbp]
\begin{center}
\includegraphics[width=1\linewidth]{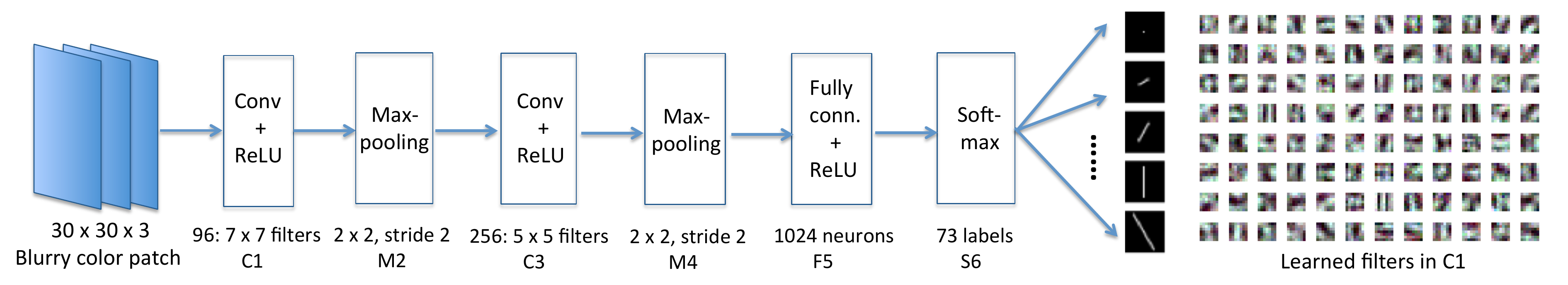}\label{fig:network}
\end{center}
\vspace{-0.4cm}
   \caption{Structure of CNN for motion kernels prediction. It is composed of 6 layers of convolutional layers and fully connected layers. It outputs the probability of each candidate motion kernel using soft-max layer. The right sub-figure shows the learned filters in C1.}
\label{fig:networkStructures}
\vspace{-0.4cm}
\end{figure*}

We now present our approach to predicting motion blur kernels (or equivalently, the motion vector) at the patch level. We decompose the image into overlapping patches of size $30 \times 30$. Given a blurry patch $\Psi_p$ centered at pixel $p$, we aim to predict the probabilistic distribution of motion kernels: 
\vspace{-0.1cm}
\begin{equation}\label{eqn:motionDist}
P(\mathbf{m} = (l, o)|\Psi_p)
\end{equation}
for all $\l \in S^{l}$ and $o \in S^{o}$, $S^{l}$ and $S^{o}$ are the sets of motion lengths and orientations respectively.  In the followings, we call this distribution as \textit{\textbf{motion distribution}}. 

Taking the problem of motion kernel estimation as a learning problem,  we utilize convolutional neural network to learn the effective features for predicting motion distributions in Eqn.~(\ref{eqn:motionDist}).  We generate a set of candidate motion kernels by discretizing the motion space, i.e., the ranges of length and orientation of the motion vectors. In our implementation, we discretize the range of motion length into 13 samples from $l = 1$ to $25$ with interval of two, and discretize the range of motion orientation $[0, 180^\circ)$ into 6 samples from $ 0^{\circ}$ to $150^{\circ}$ with interval of $30^{\circ}$. Note that when the motion length $l = 1$, all motion vectors correspond to the same blur kernel (i.e., identity kernel) on image grid regardless of the motion orientation. We therefore generate $73$ candidate motion vectors (shown in Fig.~\ref{fig:motion_discretize}(c)) in different combinations of motion lengths and orientations.  We denote the above set of motion kernel candidates as $S$ and the sets of motion lengths and motion orientations as $S^{l}$ and $S^{o}$ respectively. Obviously, these candidate motion vectors are far from dense in the continuous motion space. In Section~\ref{sec:lext} we will show how to extend the motion kernels of CNN to predict motion kernels outside the set $S$.

Given the candidate motion kernel set $S$, we next construct and learn CNN for predicting the motion distribution over $S$ given a blurry patch.  The convolutional neural network is constructed as follows. As shown in Fig.~\ref{fig:networkStructures}, the network has  six layers: $C1-M2-C3-M4-F5-S6$.
$C1$ is a convolutional layer using filters ($7 \times 7 \times 3$) followed by ReLU (i.e., $f(x) = max(x, 0)$~\cite{krizhevsky2012imagenet}) non-linear transform; 
$M2$ is a max-pooling layer over $2 \times 2$ cells with stride 2;  $C3$ is a convolutional layer using 256 filters ($5 \times 5 \times 96$); $M4$ is a max-pooling layer same as $M2$; $F5$ is a fully connected layer with 1024 neurons; $S6$ is a soft-max layer with 73 labels, and each label corresponds to a candidate motion blur kernel in $S$ as  shown in Fig.~\ref{fig:motion_discretize}(c).

To train the CNN model, we generate a large set of training data $T = \{\Psi_k, \mathbf{m}_k\}_{k=1}^K$, which are composed of  blurry patch / motion kernel pairs. We synthetically generate blurry images by convolving clean natural images with the 73 possible motion kernels, then randomly crop $30 \times 30 \times 3$ color patches from the blurry images as the training patches $\{\Psi_k\}_{k=1}^K$, and take the labels of corresponding ground-truth motion kernels as the training labels $\{\mathbf{m}_k\}_{k=1}^K$. We generate training data using 1000 images randomly sampled from PASCAL VOC 2010 database and finally construct a training set of around 1.4 million pairs of blurry patches and their ground-truth motion kernels. 
Using Caffe~\cite{donahue2013decaf}\footnote{http://caffe.berkeleyvision.org}, we train the CNN model in one million iterations by stochastic gradient descent algorithm with batches of 64 patches in each iteration. 

Because the final layer of the CNN is a soft-max layer, we can predict the probabilities of motion kernels given an observed blurry patch $\Psi$ as 
\vspace{-0.2cm}
\begin{equation}
P(\mathbf{m} =(l, o)|\Psi) = \frac{\exp((w^{S6}_{c})^T \phi_{F5}(\Psi))}{\sum_{n}\exp((w^{S6}_{n})^T \phi_{F5}(\Psi))},
\end{equation}
where $w^{S6}_c$ is the vector of weights on neuron connections from F5 layer to the neuron in $S6$ layer representing the motion kernel $(l, o)$, $c$ is the index of $(l, o)$ in $S$. $\phi_{F5}(\Psi)$ is the output features of $F5$ layer of a blurry patch $\Psi$, which is a 1024-dimensional feature vector.


In our implementation, we also tried to learn more complex CNN structures (e.g., with one more convolutional layer or more filters in convolutional layers),  but the learning speed is significantly slower while the final prediction results are not significantly improved. Figure~\ref{fig:networkStructures} (right) shows examples of automatically learned filters by our CNN model for motion kernel prediction. These filters reflect diverse local structures in sharp or blurry patch instances.

\begin{figure}[htbp]
\begin{center}
\includegraphics[width=0.95\linewidth]{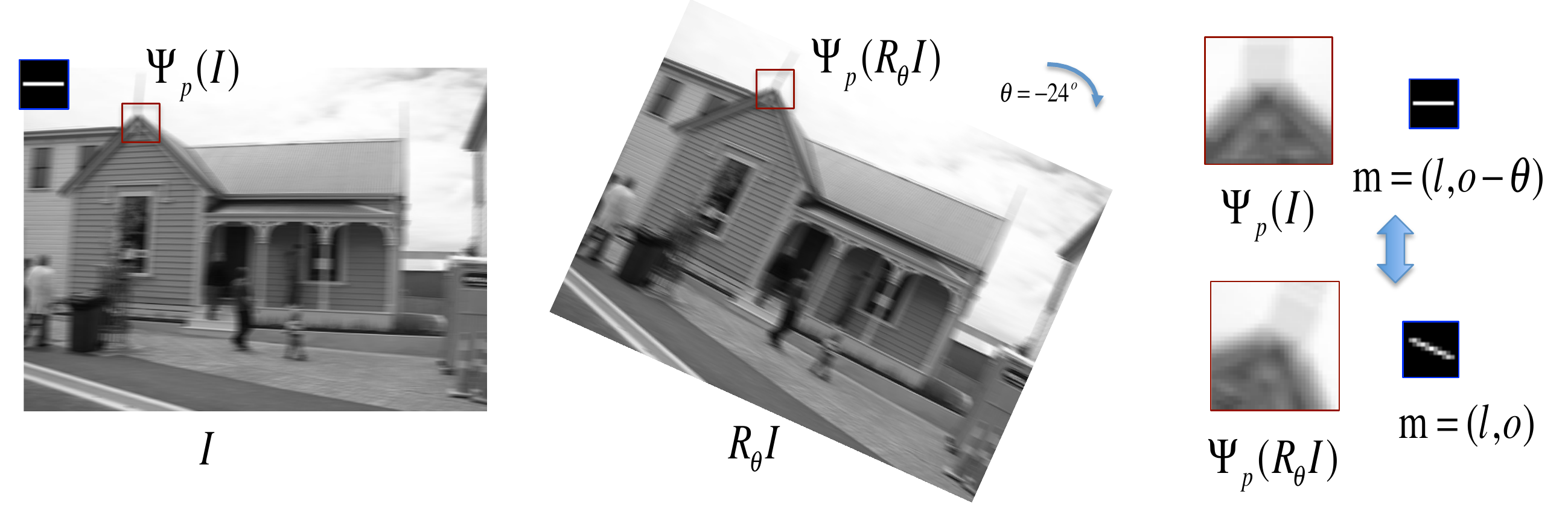}
\end{center}
\vspace{-0.3cm}
   \caption{Motion kernel estimation on a rotated patch. $I$ is a blurry image, $R_{\theta} I$ is the rotated image with~$\theta$ ($\theta = -24^o$ in this case).}
\label{fig:extension}
\vspace{-0.5cm}
\end{figure}

\begin{figure*}[htbp]
\begin{center}
\includegraphics[width=0.96\linewidth]{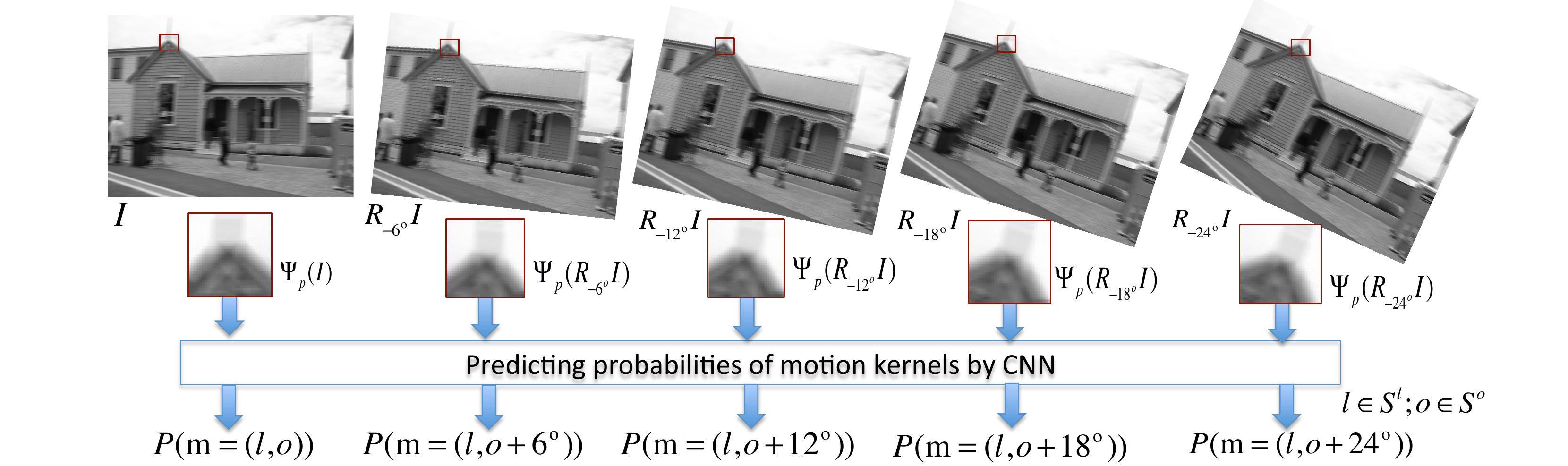}
\vspace{-0.3cm}
\end{center}
   \caption{Extension of motion kernel set predicted by CNN using rotated images. For an image $I$,  we generate its rotated images $R_{-6^{\circ}}I, R_{-12^{\circ}}I, R_{-18^{\circ}}I, R_{-24^{\circ}}I$, then feed each patch and its rotated versions into the CNN to predict  motion distributions.  By concatenating all the motion distribution estimations, we can estimate the probabilities of more densely sampled motion kernels.}
\label{fig:extension_p2}
\vspace{-0.5cm}
\end{figure*}

\subsection{Extending the Motion Kernel Set of CNN}~\label{sec:lext}
\vspace{-0.05cm}
Our learned CNN model can predict the probabilities of 73 candidate motion kernels in $S$.  Obviously, they are not sufficiently dense in the motion space. We next extend the motion kernel set predicted by the CNN to  enable the prediction for  motion kernels outside $S$.



We make the extension based on the following observation. As shown in Fig.~\ref{fig:extension},  given a blurry image $I$, we rotate it by $\theta$ degrees (denoted as $R_{\theta} I$, $R_{\theta}$ is a rotation operator). For a pair of patches $\Psi_p(I)$ and its rotated version $\Psi_p(R_{\theta}I)$ cropped from $I$ and $R_{\theta}I$ centered at pixel $p$ respectively, if we can predict that  the motion kernel of $\Psi_p(R_{\theta}I)$ is $\mathbf{m} = (l, o)$, then we can deduce directly that the motion kernel of corresponding patch $\Psi_p(I)$ in $I$ is  $\mathbf{m} = (l, o - \theta)$. 

Based on the above observation, we can estimate the probabilities of motion kernels for patch $\Psi_p(I)$ using its rotated patch $\Psi_p(R_{\theta}I)$. By feeding the rotated patch into CNN, we can estimate the probabilities of motion kernels for the rotated patch: $P(\mathbf{m} = (l, o)| \Psi_p(R_{\theta}I)), \mathbf{m} \in S$, then we can deduce that the motion distribution of the original patch $\Psi_p(I)$ before rotation is:
\begin{eqnarray} \label{eqn:extend1}
P(\mathbf{m} = (l, o - \theta) | \Psi_p(I)) = P(\mathbf{m} = (l, o) | \Psi_p(R_{\theta}I)).
\end{eqnarray} 
Note that motion $\mathbf{m} = (l, o - \theta)$ may not belong to the motion kernel set of CNN (i.e., $S$).


By carefully designing the image rotations, we can extend the motion kernel set of CNN as follows. Remember that the original CNN can predict probabilities of 73 motion kernels in $S$ with orientations in $S^{o} = \{0^\circ , 30^\circ, 60^\circ, 90^\circ, 120^\circ, 150^\circ \}$ with interval of $30^\circ$.  As shown in Fig.~\ref{fig:extension_p2},  given a blurry image $I$,  we first generate its rotated images $R_{-6^{\circ}}I, R_{-12^{\circ}}I, R_{-18^{\circ}}I, R_{-24^{\circ}}I$ with rotation angles within $[0, 30^{\circ})$ and interval of $6^{\circ}$. For each patch $\Psi_p(I)$ centered at pixel $p$, we extract its rotated versions $\Psi_p(R_{\theta}I)$ ($\theta \in \{-6^{\circ}, -12^{\circ}, -18^{\circ}, -24^{\circ}\}$) from the rotated images. By feeding these patches  into CNN, we can predict the probabilities of motion kernels for patch $\Psi_p(I)$ using each patch based on Eqn.~(\ref{eqn:extend1}):
\begin{eqnarray}
\nonumber P(\mathbf{m} = (l, o) | \Psi_p(I)) = P(\mathbf{m} = (l, o) | \Psi_p(I)), \\
\nonumber P(\mathbf{m} = (l, o + 6 ^ \circ) | \Psi_p(I)) = P(\mathbf{m} = (l, o) | \Psi_p(R_{-6 ^ \circ}I)) ,\\\nonumber P(\mathbf{m} = (l, o + 12 ^ \circ) | \Psi_p(I)) = P(\mathbf{m} = (l, o) | \Psi_p(R_{-12 ^ \circ}I)), \\ \nonumber P(\mathbf{m} = (l, o + 18 ^ \circ) | \Psi_p(I)) = P(\mathbf{m} = (l, o) | \Psi_p(R_{-18 ^ \circ}I)), \\
\nonumber P(\mathbf{m} = (l, o + 24 ^ \circ) | \Psi_p(I)) = P(\mathbf{m} = (l, o) | \Psi_p(R_{-24 ^ \circ}I)),
\end{eqnarray}
where $l \in S^{l}, o \in S^{o}$. By concatenating all the above estimations from one patch and its rotated versions, we can therefore predict the motion distribution in an extended motion kernel set of CNN: $P(\mathbf{m} = (l, o) | \Psi_p(I))$, $o \in S_{ext}^{o} = \{0^\circ, 6^\circ, 12^\circ, \cdots, 174^\circ\}, l \in {S^{l}}$.
After motion kernels extension, we can totally predict probabilities of 361 candidate motion kernels\footnote{There are totally 390 possible kernels by combining the motion lengths in $S^{l}$ ($|S^{l}| = 13$) and motion orientations in $S^{o}_{ext}$  ($|S^{o}_{ext}| = 30$). But the motion kernels $\{\mathbf{m} = (l, o)\}_{o \in S^{o}_{ext}}$ when $l = 1$ are all the same, we only retain one of them. } for an image patch by CNN, which is almost 5 times of the number of candidate motion kernels in $S$ predicted by CNN. Note that this process does not require  the CNN retraining, but just feed this image and its rotated versions to our learned CNN. 

\begin{figure}[htbp] 
\begin{center}
\includegraphics[width=1.0\linewidth]{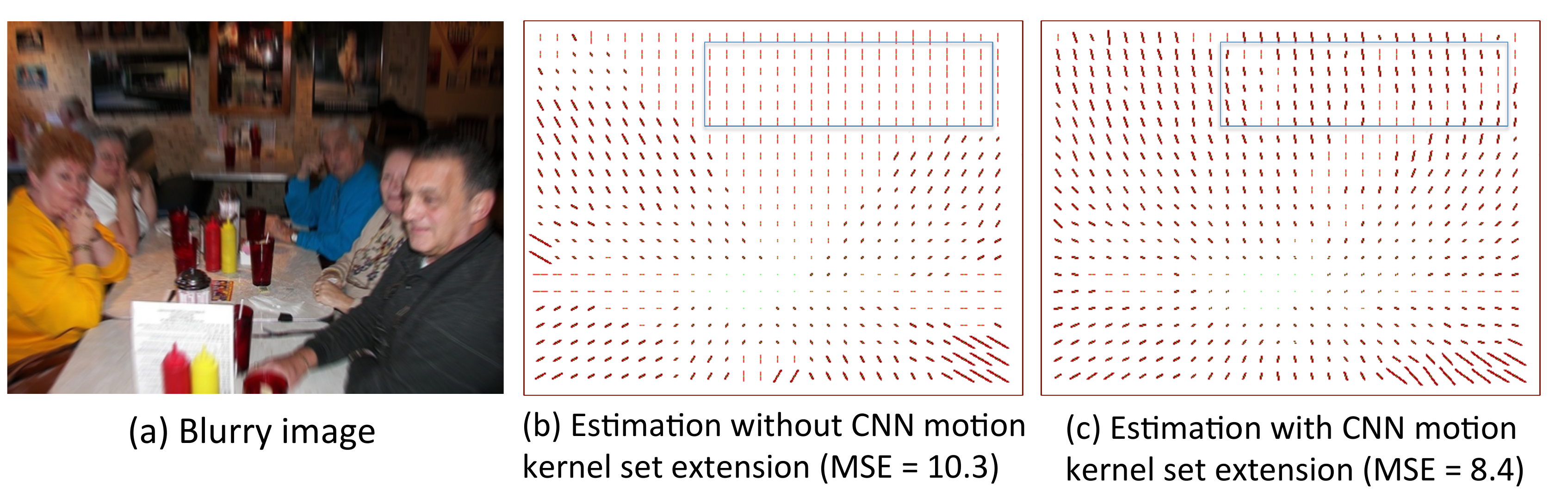}
\end{center}
\vspace{-0.3cm}
   \caption{Effect of CNN motion kernel set extension. }
\label{fig:LabelExtent}
\vspace{-0.3cm}
\end{figure}
Figure~\ref{fig:LabelExtent} shows an example of  motion kernel estimation without and with CNN motion kernel set extension. In this example, we synthetically generate the motion blur using a camera motion. As shown in Fig.~\ref{fig:LabelExtent}(b), the estimated motion kernels suffer from blocky artifacts in the blue rectangle because all pixels in it are predicted to have the same orientation due to the large quantization interval of motion orientations. By extending the motion kernel set of CNN, we can predict more accurate motion kernels shown in Fig.~\ref{fig:LabelExtent}(c). The mean squared error (MSE) w.r.t. ground-truth motion kernels is reduced from 10.3 to 8.4. 
\begin{figure*}[htbp] \label{fig:ex_confMap}
\begin{center}
\includegraphics[width=1.0\linewidth]{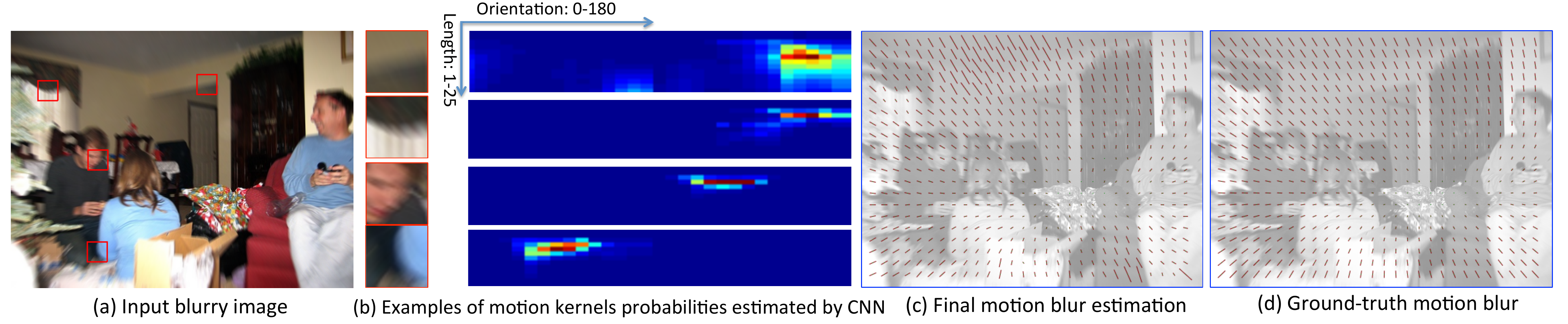}
\vspace{-0.7cm}
\end{center}
   \caption{Examples of motion kernel probabilities. The left of (b) show four blurry patches cropped from (a). Each color map on the right of (b) shows the probabilities of motion kernels in different motion lengths and orientations estimated for each blurry patch by CNN. Note that the high probability regions are local in each map. (c) shows our final motion kernel estimation.}
\label{fig:ex_confMap}
\vspace{-0.2cm}
\end{figure*}



\begin{figure*}[!htbp] 
\begin{center}
\includegraphics[width=1.0\linewidth]{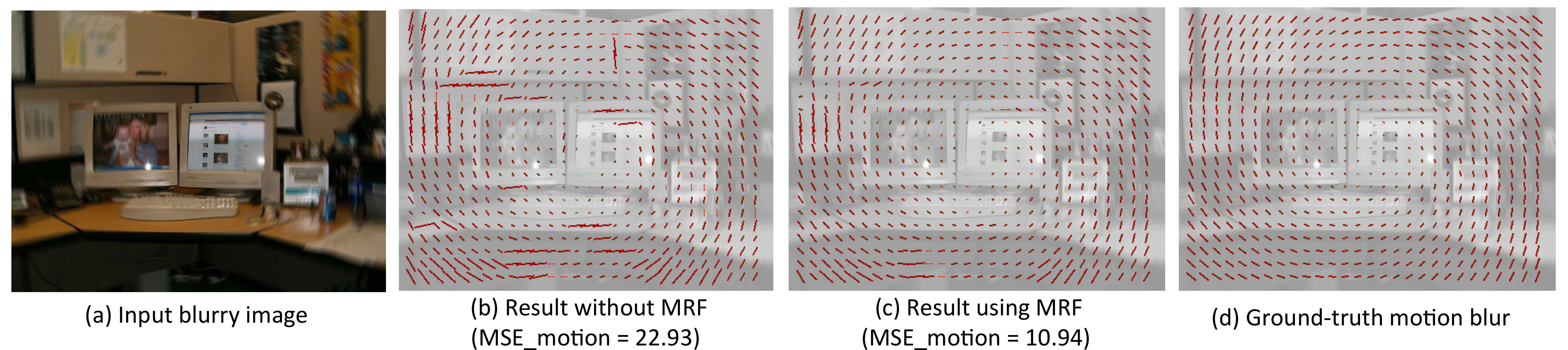}
\end{center}
\vspace{-0.4cm}
   \caption{Example of non-uniform motion kernel estimation. (b) Estimation using the unary term of Eqn.(\ref{eqn:mrf}), i.e., choosing the motion kernel with highest confidence for each pixel. (c) Estimation using the full model of Eqn.(\ref{eqn:mrf}) with motion smoothness constraint. (d) Ground-truth motion blur. MSE\_motion is an accuracy measurement of motion blur defined in Section 5. }
\label{fig:ex_MRF}
\vspace{-0.4cm}
\end{figure*}

\vspace{-0.2cm}
\section{Dense Motion Field Estimation by MRF} \label{sec:denseMotion}
The CNN predicts distribution of motion kernels in an image at the patch level. We now discuss how to fuse these patch-level motion kernel estimations into a dense field of motion kernels for the image. 

Given an image $I$, we sample $30 \times 30 \times 3$ overlapping color patches with a spatial interval of  $6$ pixels over image $I$. Each patch $\Psi_p(I)$ produces motion kernel probabilities: $P(\mathbf{m} = (l, o)| \Psi_p(I))$ ($l \in S^{l}, o \in S^{o}_{ext}$) by applying the CNN.  Figure~\ref{fig:ex_confMap}(b) shows examples of motion distribution maps for four blurry patches, and each of them is sparse and composed of one local high probability region.  We assume that the pixels in patch $\Psi_p(I)$ share the same motion distribution. Then each pixel has multiple estimates for each motion kernel probability from all patches containing it. For a pixel $p$, we perform weighted average over the multiple estimates of each motion kernel probability, and define the confidence of motion kernel $\mathbf{m} = (l, o)$ at pixel $p$ as
\begin{eqnarray}
\nonumber C(\mathbf{m}_p = (l, o)) = \\ \frac{1}{Z} \sum_{q: p \in \Psi_q} G_\sigma (||x_p - x_q ||^2)P(\mathbf{m} = (l, o) | \Psi_q),
\end{eqnarray}
for all $l \in S^{l}, o \in S^{o}_{ext}$. $x_p$ is the coordinate of pixel $p$. As a Gaussian function, $G_\sigma (||x_p - x_q ||^2)$ imposes higher weights on the patch $\Psi_q$ in the summation if its center pixel $q$ is closer to pixel $p$. $\sigma$ is set to 10 in our implementation. This means that we trust more the motion prediction from the patch containing pixel $p$ closer to its patch center. $Z = \sum_{q: p \in \Psi_q} G_\sigma (||x_p - x_q ||)$ is a normalization constant.


We further assume that the motion kernels are spatially smooth. This is reasonable because the moving objects or camera are moving smoothly during capturing image, and nearby pixels should have similar motions. Then we estimate the dense motion field $M = \{\mathbf{m}_p = (l_p, o_p)\}_{p \in \Omega}$ over image $I$ by optimizing the following MRF model:
\vspace{-0.2cm}
\begin{eqnarray} \label{eqn:mrf}
\nonumber {\min}_{M}\sum_{p \in \Omega} [-C(\mathbf{m}_p = (l_p, o_p))  + \\
 \sum_{q \in N(p)}{\lambda [(u_p - u_q)^2 + (v_p - v_q)^2] },
\end{eqnarray}
where $l_p\in S^{l}, o_p \in S^{o}_{ext}$, $(u_p, v_p) $ and  $(u_q, v_q) $  are motion vectors $\mathbf{m}_p$ and $\mathbf{m}_q$ in Cartesian coordinates that are related to $(l_p, o_p)$ and $(l_q, o_q)$ by Eqn.~(\ref{eqn:catesian}). $N(p)$ is the neighborhood of  $p$. By minimizing the energy function, the first term encourages to choose the motion kernel for each pixel with higher confidence estimated by CNN, and the second term enforces the smoothness of nearby motion kernels. 

For each pixel, there are 361 motion kernel candidates, it is inefficient to optimize the MRF problem with such a large number of candidate labels. We therefore generate candidate motion kernels for each pixel by selecting the top 20 motion kernels with highest confidence values, together with 30 sampled motion candidates from the remaining candidates  to make the motion kernel candidate set for each pixel both prominent and diverse.  Since the candidate label sets are spatially varying, we cannot use the off-the-shelf graph cut toolbox~\cite{szeliski2008comparative}, we therefore optimize the energy by max-product belief propagation algorithm. Predicting dense motion blur for an image of size $300 \times 400$ takes around 80 seconds using CPU including computing patch-level motion distributions by CNN.

Figure~\ref{fig:ex_MRF} shows an example of motion blur estimation. As shown in Fig.~\ref{fig:ex_MRF}(b, c), the full MRF model can effectively remove the noisy estimates in Fig.~\ref{fig:ex_MRF}(b)  using smoothness term, and quantitative results are significantly improved. 

\section{Non-Uniform Motion Deblurring}
With the dense non-uniform motion kernels estimated by CNN, we now deconvolve the blurry image to estimate the sharp image. It is challenging to deconvolve the image blurred by non-uniform motion blur. We adapt the uniform deconvolution approach in~\cite{zoran2011learning} to the non-uniform deconvolution problem. The non-uniform deconvolution is modeled as optimizing: 
\begin{equation}~\label{eqn:deblur}
{\min}_{I}\frac{\lambda}{2} || k(M) *  I - O||_2^2 - \sum_{i \in \Omega}  \log(P(R_iI))
\end{equation}
where $O$ is the observed blurry image, $R_i$ is an operator to extract the patch located at $i$ from  an image. $P(\cdot)$ is the prior distribution of natural image patches, which is modeled as a Gaussian mixture model learned from natural image patches~\cite{zoran2011learning}.  

Different to uniform deblur in~\cite{zoran2011learning}, the first term in Eqn.(\ref{eqn:deblur}) is modeled for non-uniform motion blur.
We optimize the above problem by half-quadratic splitting algorithm, i.e., optimizing: 
$
{\min}_{I,\{z_i\}}\frac{\lambda}{2} || k(M) *  I - O||_2^2  + \sum_{i \in \Omega} (\frac{\beta}{2} ||R_i I - z_i||_2^2- \log(P(z_i))),
$
where auxiliary variables $\{z_i\}$ are introduced.  We iteratively optimize $I$ and $\{z_i\}$ by increasing $\beta$. In the iterations, we need to optimize the following two sub-problems. (1) By fixing $\{z_i\}$, we optimize sharp image: 
${\min}_{I}\frac{\lambda}{2} || k(M) *  I - O||_2^2  + \sum_{i \in \Omega} (\frac{\beta}{2} ||R_i I - z_i||_2^2).$
(2) By fixing $I$, we optimize $\{z_i\}$:
$\min_{z_i} \frac{\beta}{2} ||R_i I - z_i||_2^2 -   \log(P(z_i)), i \in \Omega.$

For sub-problem (1), the blur kernels $k(M)$ are non-uniform and determined by spatially varying motion vectors $M$. By re-writing the non-uniform convolution as matrix-vector multiplication (i.e., $k(M)*I = K_MI$ ), we optimize sharp image by solving the linear equations deduced by setting the gradients of cost in sub-problem (1) to zeros:
\begin{eqnarray}\label{eqn:subprob1_solve}
[\lambda K_M^TK_M + \beta \sum_{i \in \Omega}(R_i^TR_i)] I  =  \lambda K_M ^ T O + \beta(\sum_{i \in \Omega}{ R_i^T z_i}).
\end{eqnarray}
We solve these linear equations using a conjugate gradient algorithm. In the implementation, all the involved matrix-vector multiplications can be efficiently implemented by convolutions or local operations around each pixel. $R_i^T z$ is an operation to put the patch $z$ back to the region where it was extracted.  
The sub-problem (2) can be optimized following~\cite{zoran2011learning}. In implementation, we set the patch size to $8 \times 8$, $\lambda = 2 \times 10^5$, and $\beta$ is increased from 50 to 3200 with a ratio of 2 in 7 iterations of alternative optimizations.

\vspace{-0.15cm}
\section{Experiments}

\vspace{-0.2cm}

\begin{table}
\caption {Comparison of motion kernel estimation on 15 test images with synthetic motion blur. ``BlurSpect'' is based on the approach in~\cite{chakrabarti2010analyzing}.  ``SLayerRegr'' is the extension of approach in~\cite{couzinie2013learning}.   }
\label{tab:comp_motion}
\vspace{0.1cm}
\scriptsize
\label {tab:motionEsti}
\centering 
\begin {tabular}{ |p{1.2cm}<{\centering}|p{0.9cm}<{\centering}|p{1.0cm}<{\centering}|p{0.9cm}<{\centering}|p{0.9cm}<{\centering}|p{1.0cm}<{\centering}|}
\hline
Methods & DL\_MRF & {DL\_noMRF} & DL\_noLE & BlurSpect & SLayerRegr   \\ \hline
MSE\_motion &\bf{7.83}  & 16.35 & 9.01 & 44.56 &65.10   \\
PSNR\_motion  & \bf{44.55} &  $37.14$  & 43.17 & 26.58& 22.70 \\  \hline
\end {tabular}
\vspace{-0.2cm}
\end {table}

\begin{figure*}[!htbp]
\begin{center}
\includegraphics[width=1.0\linewidth]{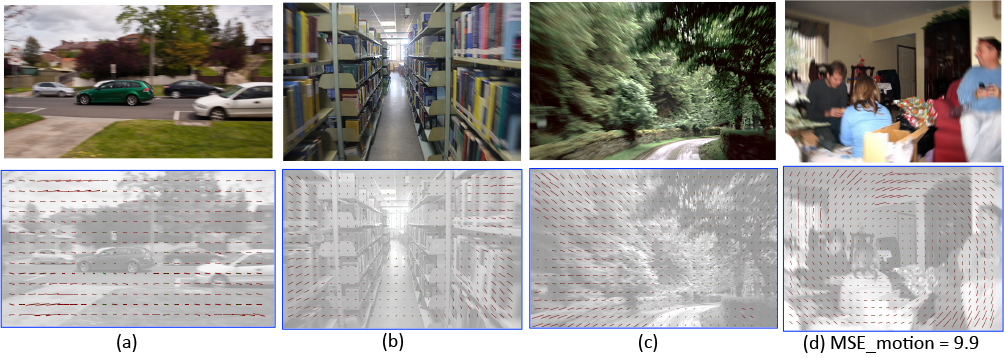}
\vspace{-0.8cm}
\end{center}
   \caption{Examples on motion kernel estimation.  The first three columns are real blurry images, the last column shows a synthetic picture with camera rotation (MSE\_motion = 9.9). }
\label{fig:motionComp}
\vspace{-0.1cm}
\end{figure*}

To evaluate the quantitative accuracy of our approach for non-uniform motion kernel estimation, we generate 15 synthetic blurred images with ground-truth non-uniform motion kernels caused by camera motions (rotation and translation). The examples shown in Figs.~\ref{fig:extension_p2}-\ref{fig:ex_MRF} are from this synthetic image set. Given the estimated motion blur kernels $M = \{u_p, v_p\}_{p \in \Omega}$ and ground-truth motion blur kernels $M^{gt} = \{u^{gt}_p,  v^{gt}_p\}_{p \in \Omega}$ in the Cartesian coordinate system, we measure the accuracy of the estimated motion kernel by the mean-squared-error (MSE\_motion): $\mathbf{MSE\_{motion}}(M, M^{gt}) = \frac{1}{2 |\Omega|}\sum_{p \in \Omega}[(u_p - u_p^{gt})^2 + (v_p - v^{gt}_p)^2]$ and peak signal-to-noise ratio (PSNR\_motion): $\mathbf{PSNR\_{motion}}(M, M^{gt}) = -10 \log{\frac{\mathbf{MSE\_motion}(M, M^{gt})}{d^2_{max}}}$, $d_{max} = 25$ is the maximum motion length.

\begin{figure*}[htbp]
\begin{center}
\includegraphics[width=0.95\linewidth]{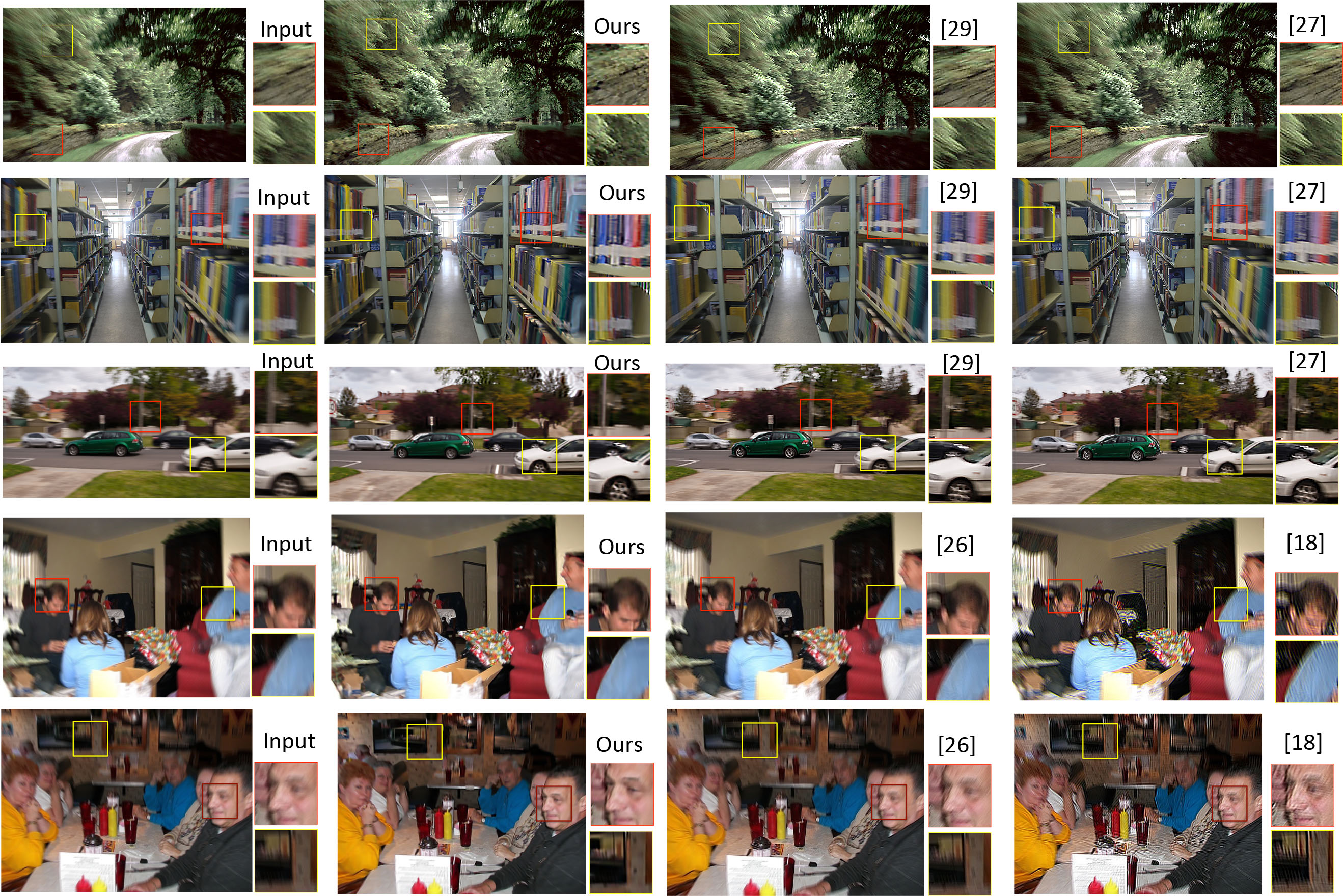}
\vspace{-0.3cm}
\end{center}
   \caption{Examples of non-uniform motion deblurring. The first and second columns show the blurry images and our results.  The third and fourth columns show the results of methods in~\cite{Levin,whyte2012,xu2010two,xuL0} using their source codes. These examples are challenging because the motion blur kernels are strongly non-uniform and the scenes are complex. Our estimated motion blur fields are shown in Figs.~\ref{fig:motionComp}, \ref{fig:motionExam} .}
\label{fig:exSet_deblur}
\vspace{-0.2cm}
\end{figure*}

\begin{table*}[!htbp]
\caption {Accuracies of  motion kernel estimation and blur removal  on 15 test images with synthetic motion blur. ``BlurSpect'' is based on the approach in~\cite{chakrabarti2010analyzing}.  ``SLayerRegr'' is the extension of approach in~\cite{couzinie2013learning}. ``MSE\_ker'' is an error for
non-uniform blur kernel estimation using the average MSE of blur kernels across image pixels. ``PSNR\_deblur'' is the PSNR of the final deblurred results. The number in each table cell is the mean value over the image set. }
\label{tab:comp_motionDeblur}
\vspace{0.1cm}
\scriptsize
\label {tab:motionEsti}
\centering 
\begin {tabular}{ |p{1.2cm}<{\centering}|p{1.0cm}<{\centering}|p{1.0cm}<{\centering}|p{1.0cm}<{\centering}|p{1.4cm}<{\centering}|p{1.4cm}<{\centering}|p{1.4cm}<{\centering}|p{1.4cm}<{\centering}|p{1.4cm}<{\centering}|p{1.4cm}<{\centering}| }
\hline
 &DL\_MRF & {DL\_noMRF} & DL\_noLE & BlurSpect & SLayerRegr  & TwoPhase~\cite{xu2010two}& MargLike~\cite{Levin} & NonUnif~\cite{whyte2012} & UnNatural~\cite{xuL0}\\ \hline
MSE\_ker & \bf 0.024 &0.029 & 0.041 & 0.250 & 0.127 & 0.108&0.119 &0.193 &0.165\\
PSNR\_deblur& \bf 24.81 & 24.66&  $24.61$  & 21.72 & 19.04 &21.26 &18.49 &20.65 & 21.33\\  \hline
\end {tabular}
\end {table*}

\begin{figure*}[htbp]
\begin{center}
\includegraphics[width=0.95\linewidth]{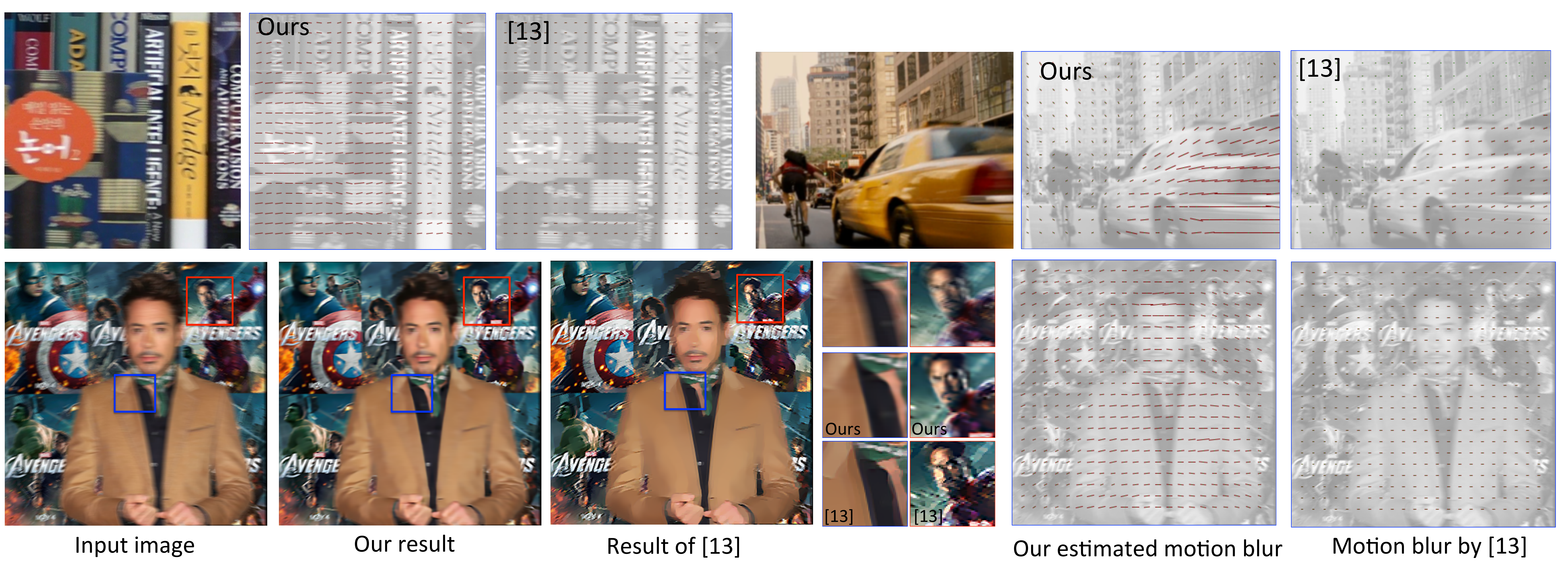}
\vspace{-0.4cm}
\end{center}
   \caption{Comparison to~\cite{kim2014segmentation}. Our CNN can better predict the different motion layers.  The deblurring result of~\cite{kim2014segmentation} is over-sharpened and image details are removed, while our result is visually more natural. }
\label{fig:deblur_compare_kim}
\end{figure*}

\begin{figure*}[!htbp]
\begin{center}
\includegraphics[width=1.0\linewidth]{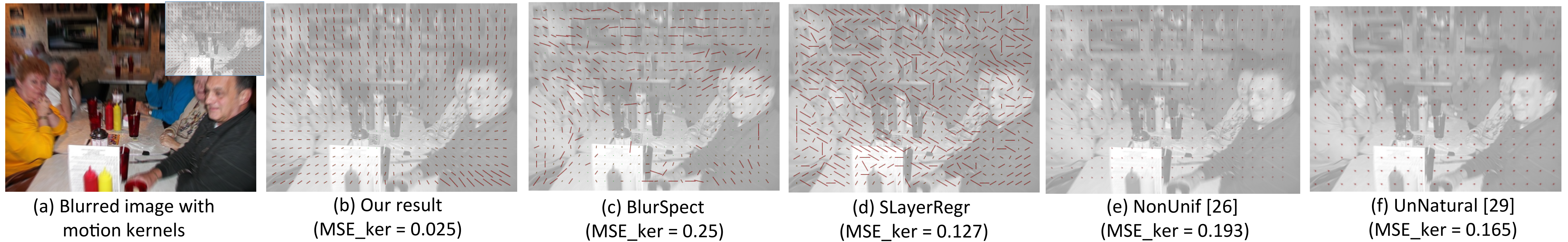}
\vspace{-0.7cm}
\end{center}
   \caption{Comparison of motion kernel estimation. ``BlurSpect'' and ``SLayerRegr'' are based on the methods in~\cite{chakrabarti2010analyzing} and~\cite{couzinie2013learning}  respectively.   }
\label{fig:motionExam}
\vspace{-0.3cm}
\end{figure*}
Figure~\ref{fig:motionComp}  presents four examples with strongly non-uniform motion blur captured for scenes with complex depth layers. The first three examples are real-captured blurry images, and the final example is a synthetic blurry image. All these examples show that our CNN-based approach can effectively predict the spatially varying motion kernels.

In Table 1, we evaluate and compare our approach to the other approaches for non-uniform motion kernel estimation. ``DL\_noMRF'' is our approach using only the unary term in Eqn.~(\ref{eqn:mrf}). ``DL\_noLE'' is our MRF-based approach without using the motion kernel set extension. ``DL\_MRF'' is our full estimation approach. ``BlurSpect''  is the approach proposed in~\cite{chakrabarti2010analyzing}.  It was originally designed for estimating horizontal or vertical motion blur, and we extend it to estimate motion kernels with orientations in $S^{o}_{ext}$ by the technique in Section~\ref{sec:lext}. ``SLayerRegr'' is an extended version of approach in~\cite{couzinie2013learning}. The original approach learns a logistic regressor to estimate discrete motion kernels in horizontal direction using hand-crafted features. To predict motion kernels in other directions, we implement~\cite{couzinie2013learning} using the same features and learn SVMs for predicting 73 motion kernels in $S$, then extend motion kernel set by the method in Section~\ref{sec:lext}. ``SLayerRegr'' can be seen as a learning machine with a single layer of hand-crafted features. As shown in Table \ref{tab:comp_motion}, both ``BlurSpect''  and ``SLayerRegr'' perform poorly on estimating the challenging non-uniform motion blur with diverse motion lengths and orientations. Our approach can effectively estimate the motion kernels with average MSE\_motion 7.83 and PSNR\_motion 44.55. Moreover, the motion kernel set extension and motion smoothness constraint significantly improve the accuracy of motion kernel estimation.

Figure~\ref{fig:exSet_deblur} compares deblurring results of our approach, non-uniform deblurring approaches~\cite{whyte2012,xuL0} and uniform deblurring approaches~\cite{Levin, xu2010two}, for which the source codes are available. Except for ours, none of these methods handles the non-uniform blur in a satisfying manner for these examples. Our approach estimates more accurate motion blur kernels, which enables us to produce better final deblurring results. The method in~\cite{kim2014segmentation} is an effective approach for motion deblurring. Because its source code is not available, we directly compare it on examples of~\cite{kim2014segmentation}  in Fig.~\ref{fig:deblur_compare_kim}. Our approach can better ``recognize''  the complex motions. The deblurring result of ~\cite{kim2014segmentation} is commonly over-sharpened, but our deblurring result is visually more natural.

In Table~\ref{tab:comp_motionDeblur}, we qualitatively compare our method to the state-of-the-art non-blind debluring approaches for both the motion blur kernel estimation and the final deblurred results. We  define an error of
``MSE\_ker''  for
non-uniform motion blur estimation using average MSE of blur kernels across pixels in an image, and the MSE of each pixel  is defined
by the mean per-element squared difference between the estimated and ground-truth kernels after aligning kernels by centers. Contrary to the ``MSE\_motion'' that measures the kernel error in the linear motion space, this error term directly measures the kernel differences in the spatial domain. We also evaluate the deblurring result by the PSNR of the deblurred image (denoted as ``PSNR\_deblur'') w.r.t. the ground-truth clean image. All the values in  Table~\ref{tab:comp_motionDeblur} are the mean values over the image set. We can not qualitatively compare to the approach in{~\cite{kim2014segmentation}} because the source codes are not available. These results clearly show that our approach can produce significantly better results both in the motion blur kernel estimation and the motion blur removal than the compared state-of-the-art approaches.

Figure~\ref{fig:motionExam} shows an example of motion blur estimation by different non-uniform blur estimation approaches.  Our approach can produce significantly better non-uniform motion blur field than the compared approaches.

\vspace{-0.2cm}
\section{Conclusion}
\vspace{-0.15cm}
In this paper, we have proposed a novel CNN-based non-uniform motion deblurring approach. 
We learn an effective CNN for estimating motion kernels from local patches. Using an MRF model, we are able to well predict the non-uniform motion blur field. This leads to state-of-the-art motion deblurring results. 
In the future, we are interested in designing a CNN for  estimating the general non-uniform blur kernels. We are also interested in designing an CNN system that can estimate and remove general non-uniform blurs in a single framework. 


\vspace{-0.1cm}
 \section*{Acknowledgement}
\vspace{-0.1cm}
This work was partially supported by NSFC projects (61472313, 11131006), the 973 program (2013CB329404) and NCET-12-0442. Ponce was supported in part by the Institut Universitaire de France and European Research Council (VideoWorld project).

{
\bibliographystyle{ieee}
\bibliography{deblur}
}

\end{document}